\documentclass{ecai}
\usepackage{times}
\usepackage{graphicx}
\usepackage{latexsym}
\usepackage{CJK}
\usepackage{color}
\usepackage{url}
\usepackage{booktabs}
\usepackage[table,xcdraw]{xcolor}
\usepackage{todonotes}

\ecaisubmission   

\begin{document}
\begin{CJK}{UTF8}{gbsn}
\title{Merging External Bilingual Pairs into Neural Machine Translation}

\author{Tao Wang$^1$, Shaohui Kuang$^1$, Deyi Xiong \institute{Soochow University,
China, email:{rgwt1234,deyi.xiong}@gmail.com,\               shao903094405@qq.com} \and 
Ant{\'o}nio Branco \institute{University of Lisbon,
Portugal, email: antonio.branco@di.fc.ul.pt} }

\maketitle
\bibliographystyle{ecai}

\begin{abstract}
  As neural machine translation (NMT) is not easily amenable to explicit correction of errors, incorporating pre-specified translations into NMT is widely regarded as a non-trivial challenge. In this paper, we propose and explore three methods to endow NMT with pre-specified bilingual pairs. 
  Instead, for instance, of modifying the beam search algorithm during decoding or making complex modifications to the attention mechanism --- mainstream approaches to tackling this challenge ---, we experiment with the training data being appropriately pre-processed to add information about pre-specified translations. Extra embeddings are also used to distinguish pre-specified tokens from the other tokens.
  Extensive experimentation and analysis indicate that over 99\% of the pre-specified phrases are successfully translated (given a 85\% baseline) and that there is also a substantive improvement in translation quality with the methods explored here.
\end{abstract}

\section{Introduction}


As neural machine translation (NMT) is becoming the mainstream approach to machine translation, its end-to-end training makes the translation process hardly amenable to explicitly integrate principled measures to correct detected errors. 

There are many usage scenarios where it is desirable that NMT systems employ pre-specified translations from an external database. For example, in multilingual e-commerce, many brand names of products are unambiguous and have straightforward translations into the target language. Incorrect translations of such brand names will result in trade disputes. 

In general, given a source sentence with words $w_i$, $s = w_1, w_2, ..., w_n$, where there occurs the source expression $p$ of a translation pair ($p = w_k, ..., w_l, q$) that happens to be stored in the external database, $p$ should be directly translated into $q$ while other expressions in $s$ should be translated as a result of the regular running of the NMT system. 
This raises non-trivial challenges. On the one hand, NMT operates in an underlying continuous real-valued space rather than in a discrete symbolic space of pre-specified translations. On the other hand, NMT produces target translations in a word-by-word generation fashion while pre-specified translations generally contain pairs of multi-word expressions.

In this paper, we explore three methods, which we term tagging, mixed phrase and extra embedding methods, to endow NMT with pre-specified translations. In a pre-processing stage, the beginning and the end of the textual occurrences of expressions from an external database are marked with special purpose tags. Additionally, the source expressions are added with their translational equivalent target expressions. And an extra embedding is also applied to distinguish pre-specified phrases from the other words. 

We conduct experiments with both Chinese-to-English and English-to-German NMT in the news domain. On Chinese-to-English translation, by applying a combination of these three methods, the probability of successfully translating increases from 74.19\%, with the baseline, to 98.40\%, with our proposed methods, at the sentence level, and from 85.03\% to 99.13\%, at the phrase level. This approach also achieves a competitive improvement of 1.4 BLEU points over a baseline of 45.7 points. On English-to-German translation, the probability of successfully translating at the phrase level is also improved to 99.54\% from 93.69\% (baseline) with our methods. 

\section{NMT with Pre-Specified Translations}
\label{2}

The challenge in incorporating an external pre-specified translation pair $(p, q)$ into NMT, where the expression $p$ is present in the source sentence, is that $p$ should be translated or copied as the expression $q$ into a consecutive sequence of words that form the target sentence 
while the other words in the source sentence are translated by the general NMT process.

To address this challenge, we propose to experiment with three mechanisms, and their combinations, which we term tagging, mixed phrase and extra embeddings methods, described below.  The first two do not require any change in the NMT network architecture nor in the decoding algorithm.

\subsection{Methods}\label{2.1}

\textbf{Tagging}
The first method is quite straightforward. In the training data set, both the source-side phrase $p$ and its target equivalent $q$ are each surrounded with two markers, namely $<$\texttt{start}$>$ and $<$\texttt{end}$>$. As the NMT engine will be trained, these tags will have their own word embeddings automatically learned, just like any other words in the source and target sentences. This will permit to build a connection between $p$ and $q$ because of the same pattern where they occur.
This connection becomes strong when we use shared embeddings because it help NMT learn the correspondence between $p$ and $q$ with stronger evidence.

``我 爱 $<$start$>$ 香港 $<$end$>$ $|$ i love $<$start$>$ hong kong $<$end$>$'' is an example of a pair of aligned sentences whose named entities (NEs) for ``hong kong'' were tagged.

\noindent\textbf{Mixed phrase}
It is expected to be far easier for a deep neural model to learn to copy than to translate. So we propose to extend $p$ with $q$. 

``我 爱 香港 hong kong $|$ i love hong kong'' is an example of a pair of aligned sentences whose NE for hong kong in the source side, namely 香港, was added with its target-side equivalent. This method is similar to \cite{song2019code}, but instead of just replacing source phrase with target phrase, we use a ``mixed'' source and target phrase.

This mixed phrase method can be used alone or
it can be combined with the tagging method. In such case, there will be a third tag, namely $<$\texttt{middle}$>$ marking the separation of $p$ from $q$ in the mixed phrase.

``我 爱 $<$start$>$ 香港 $<$middle$>$ hong kong $<$end$>$ $|$ i love $<$start$>$ hong kong $<$end$>$'' is an example of a pair of aligned sentences whose NEs for hong kong were mixed and tagged.

In our experiments we will be using both approaches, i.e. the mixed phrase method alone or also combined with the tagging method. 

By introducing many tagging instances of this sort in the training data, the NMT model is expected to learn a pattern that an expression enclosed in these tags should be translated as a copy of part of it.
As mentioned above, copy is much easier than translate, so we reconstruct training corpus by replacing source phrase with mixed phrase.  But instead of just replacing source phrase with target phrase, we use mixed phrase combined with tagging. We tend to add more information to source sentence rather than replacing because it may drop some important information, especially those occur less frequently. It's serious when existing error during replacing.

A key question for this type of approach is what the model learns and how it works? We will analyze it below in Section \ref{4}.

\noindent\textbf{Extra embeddings}
A third method does not involve enriching the training data with information about bilingual NE correspondences, but slightly enhancing the NMT model. For a given input token, its composed representation is constructed by summing the corresponding word embedding, its positional embedding~\cite{vaswani2017attention} and what we term extra embedding. 

For instance, in this sentence ``我 爱 $<$start$>$ 香港 $<$middle$>$ hong kong $<$end$>$'', which is taken as containing eight tokens, extra embeddings are supported by the input sequence ``\texttt{n\ n\ n\ s\ n\ t\ \ t\ n}'' in which \texttt{s} and \texttt{t} are aligned with the terms in the pre-specified translation pair while \texttt{n} with the other tokens.

Extra embeddings are used to differentiate source phrase, its pre-specified translation and other normal tokens. 
This idea comes from BERT \cite{devlin2018bert}, where sentence embeddings A and B are used to differentiate the sentences packed together in a single sequence.

Similar to what is intended with the tagging method, using extra embeddings also represent a copy signal: tokens aligned with the target-side extra embedding for \texttt{t} will tend to be ``copied'' to the target side, while the other tokens, aligned with \texttt{n} and \texttt{s}, tend to be translated ``normally''. It is of note though that using extra embeddings is a softer method when compared to tagging because the first directly integrates information into input representations without changing the training text, as it is necessary to do for the latter. 


\subsection{Annotation of the Data}\label{2.2}

The first two methods above are easy to be performed in a pre-processing step that precedes training and decoding.

Given some training data $T$ and an external bilingual database $K={(p_i,q_i)}_{i=1}^{N}$, for each source sentence $s$ in $T$, and for each $n$-gram phrase in $s$ (we set $n<6$), we look up $K$ to find if it includes a source phrase $p$ in $s$. The equivalent phrase $q$ in the corresponding target sentence will be detected by the same way if it exists. 

For instance, for the combination of the tagging and mixed phrase methods, if $(p, q)$ is detected, the matched source phrase $p$ is replaced by  $<$\texttt{start}$>$ $p$ $<$\texttt{middle}$>$ $q$ $<$\texttt{end}$>$.


After the annotation of the training data $T$ in this pre-processing step, the NMT model is trained under the usual procedure. 

The annotation of a test set, in turn, is identical to the annotation of the training set except that tagging and mixed phrase replacement is performed only on the source sentences. The test data set resulting from this annotation process is then translated by the NMT system.

\subsection{Gathering a Bilingual Database}\label{2.3}

The external bilingual database can be either automatically extracted from parallel corpora or developed by human experts. In the work reported here, it was extracted from parallel corpora. In this section, we briefly describe the procedure followed.

We focused on named entities (NEs) \cite{nadeau2007survey} and used LTP tools \cite{che2010ltp} to perform named entity recognition (NER) on the Chinese corpus. We got a phrase table using Moses \cite{koehn2007moses}. 

The translations of the recognized NEs in the phrase table are detected in the target sentences and added to the candidate list. If eventually there are multiple target options in the candidate list for a given source NE, it is selected the one with the highest probability. Detected source NEs and their translations are paired and stored in the database.

With this procedure, other types of bilingual phrases can also be automatically extracted to construct and expand the external database of pre-specified translations.

Notably, we actually used the word alignment tool in preliminary experiments. However, word alignments are with many errors especially for multi-word entities. The phrase-table provides probabilistic measures to evaluate the quality. 

\section{Experiments}\label{3}

\subsection{Base Model}\label{3.1}

\begin{figure}[t]
\centering
\includegraphics[scale=0.45]{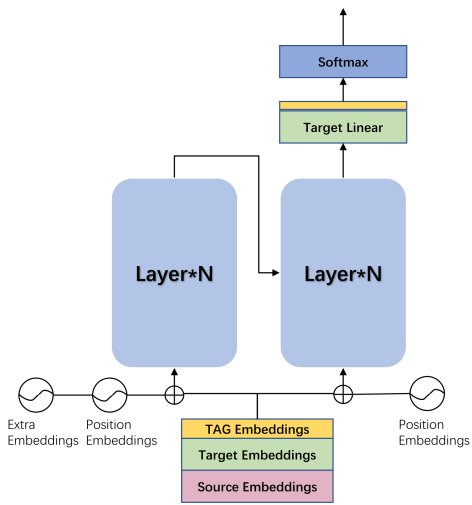}
\caption{Model architecture with shared embeddings and extra embeddings.}
\label{model architecture}
\end{figure}

To undertake our experiments, we use the Transformer model that is currently the most competitive one \cite{vaswani2017attention}, which is based on self-attention mechanism as our baseline. Given an input sequence $x_1,...,x_n$, Transformer encodes it into a sequence of continuous representations and then generates an output sequence $y_1,...,y_m$ of symbols, one element at a time.
We then introduce slight changes to this native Transformer model in order for it to accommodate our experiments. As depicted in Fig. \ref{model architecture}, the changes concern mainly two aspects.

\begin{table*}
\begin{center}
\caption{Accuracy of Chinese-English translation experiments (lines). For the development and each test set (columns NIST03-6), there are two columns, with the success rate of translating pre-specified translations at sentence level and at phrase level. If all the phrases in a sentence are translated successfully, the sentence is considered to be translated successfully. ``Total'' indicates the number of sentences containing pre-specified translations and the number of phrases with pre-specified translations.
}
\begin{tabular}{@{}l|cccccccccc@{}}
\toprule
\multicolumn{1}{c|}{} & \multicolumn{2}{c}{NIST06}   & \multicolumn{2}{c}{NIST03}   & \multicolumn{2}{c}{NIST04}    & \multicolumn{2}{c}{NIST05}   & \multicolumn{2}{c}{SUMMARY}                                            \\ \midrule
Total                 & 763          & 1,445          & 570          & 1,183          & 1,047          & 2,081          & 688          & 1,383          & 3,068                                          & 6,092                   \\ \midrule
BASE                  & 568          & 1,231          & 396          & 978           & 777           & 1,766          & 535          & 1,205          & 2,276 (74.19\%)                                 & 5,180 (85.03\%)          \\
+T                    & 584          & 1,245          & 395          & 972           & 769           & 1,750          & 541          & 1,217          & 2,289 (74.61\%)                                 & 5,184 (85.10\%)          \\
+R                    & 717          & 1,395          & 525          & 1,133          & 964           & 1,991          & 646          & 1,337          & 2,852 (92.96\%)                                 & 5,856 (96.13\%)          \\ \midrule
+T\&R                 & 748          & 1,429          & 561          & 1,174          & \textbf{1,033} & \textbf{2,066} & 676          & \textbf{1,371} & 3,018 (98.37\%)                                 & \textbf{6,040 (99.15\%)} \\
+T\&M                 & \textbf{753} & \textbf{1,435} & \textbf{565} & \textbf{1,178} & 1,026          & 2,058          & 674          & 1,369          & 3,018 (98.37\%)                                 & \textbf{6,040 (99.15\%)} \\ \midrule
+R\&E                 & 728          & 1,408          & 533          & 1,146          & 985           & 2,016          & 642          & 1,329          & 2,888 (94.13\%)                                 & 5,899 (96.83\%)          \\
+M\&E                 & 722          & 1,402          & 537          & 1,148          & 975           & 2,005          & 653          & 1,342          & 2,887 (94.10\%)                                 & 5,897 (96.80\%)          \\ \midrule
+T\&M\&E              & 750          & 1,432          & 560          & 1,171          & 1,032          & \textbf{2,066} & \textbf{677} & 1,370          & \textbf{3,019 (98.40\%)} & 6,039 (99.13\%)          \\ \bottomrule
\end{tabular}
\label{accuracy of copy}
\end{center}
\end{table*}

First, we use shared embeddings on the encoder and decoder, but the linear output part of the decoder is limited to the vocabulary size of the target. This is different from the general shared vocabulary method  applied to pairs of similar languages. 
The reason for this is twofold: because our source sentences contain mixed phrases, and also because we seek to enhance the effect of tagging. To achieve this, we first obtain, separately, the vocabularies of the source-side and of the target-side from the original corpus, and then concatenate them together. The first part of the resulting vocabulary is made of the target-side words, and the second part of source-side words. The words decoded are thus limited to the target-side part.

Second, we add extra embeddings on the encoder. We define its dimensions as [4,hidden layer size], that include norm token, source phrase token, target phrase token and pad token.

\subsection{Settings}\label{3.2}

We adopted the Chinese-English news domain corpus from the Linguist Data Consortium (LDC), with 1.25M sentence pairs, as our bilingual training data.\footnote{This corpus includes LDC2002E18, LDC2003E07, LDC2003E14, Hansards portion of LDC2004T07, LDC2004T08 and LDC2005T06.} We selected NIST06 as development set, and NIST03, NIST04, NIST05 as test sets. Case-insensitive BLEU-4 was used as the evaluation metric \cite{papineni2002bleu}, with scores obtained with the script ``mteval-v11b.pl''. The reason we don't use ``multi-bleu.perl'' is to be consistent with previous work trained on this dataset. This will make our baseline results comparable to theirs.

We also did additional experiments on the 4.5M WMT2017 English-German corpus. We selected newstest2014 as development set and newstest2016 as test set. We used spaCy\footnote{https://spacy.io/} to perform NER on English sentences of this WMT corpus and the script ``multi-bleu.perl'' to evaluate BLEU-4 score for English-to-German translation.

We used byte pair encoding compression algorithm (BPE) \cite{sennrich2016neural} to process all these data and restricted merge operations to a maximum of 30k. 
All the other parameters were the same of the base Transformer. 

We used 6 layers of self-attention both in the encoder and decoder. We set the dimensionality of all input and output layers to 512, and that of feed-forward layers to 2048. We employed 8 attention heads. During training, we used label smoothing \cite{szegedy2016rethinking} with value 0.1, attention dropout and residual dropout \cite{Hinton2012ImprovingNN} with a rate of 0.1 too. We used stochastic gradient descent algorithm Adam \cite{adam} to train the NMT models. ${\beta} 1$ and ${\beta} 2$ of Adam were set to 0.9 and 0.999, the learning rate was set to 0.001, and gradient norm was set to 5. Sentence pairs of similar length were batched together to take full advantage of GPU memory and each batch had roughly 4000 source and target tokens. According to past experience, the larger the batch size of the transformer, the better the translation quality. So we employed the delay update technology \cite{ott2018scaling}, and set the delay to 8, equaling to batch size of 32000 to run out a good result on a single graphics card. During decoding, we employed beam search algorithm and set the beam size to 6.

Following the procedures in Section \ref{2.3} and \ref{2.2}, 25,151 unique pre-specified translation pairs were extracted from the Chinese-English corpus, and in its 1.25M sentences, about 0.49M sentences bear at least one replacement of an NE.
About 60\% sentences in the Chinese-English test sets had at least one NE. As for English-German, only about 1/3 training and test data bear at least one NE replacement.

\subsection{Experiments}

The performance of the base model, described in Section \ref{3.1} and termed ``BASE'' in Tables~\ref{accuracy of copy} and \ref{bleu results of nist} with the results, was taken as our baseline.

As the baseline for the methods relying on some pre-processing or annotation of the data, we resorted to the replacement method of Song et al.~\cite{song2019code}, termed ``R'' in the tables with results. ``我 爱 hong kong $|$ i love hong kong'' is an example of a pair of aligned sentences where the source-side NE ``香港'' was replaced, in the source sentence, by the target-side ``hong kong''.

A number of further experiments were run by expanding the baselines with one of the methods presented above in Section \ref{2.1}, or some combination of them. In the result tables, the tagging, mixed phrase and extra embeddings methods are denoted as ``T'', ``M'' and ``E'', respectively. These tables display only the best performing models. 

\begin{table}
\begin{center}
\caption{BLEU scores of the Chinese-English translation experiments in Table~\ref{accuracy of copy}.}
\begin{tabular}{@{}l|lllll@{}}
\toprule
                 & N06         & N03         & N04         & N05         & Avg                                   \\ \midrule
BASE      & 45.55          & 45.12          & 46.36          & 45.61          & 45.66                                 \\
+T             & 45.81          & 46.49          & 46.98          & 45.48          & 46.19                                 \\ \midrule
+T\&R        & 46.09          & 46.69          & 46.70          & 46.26          & 46.43                                 \\
+T\&M      & \textbf{46.74} & 46.37          & \textbf{47.23} & 46.82          & 46.79                                 \\ \midrule
+R\&E        & 46.21          & 46.61          & 46.91          & 46.48          & 46.55                                 \\
+M\&E      & 46.52          & 46.81          & 47.07          & 46.72          & 46.78                                 \\ \midrule
+T\&M\&E & 46.38          & \textbf{47.33} & 47.18          & \textbf{47.34} & {\color[HTML]{333333} \textbf{47.06}} \\ \bottomrule
\end{tabular}
\label{bleu results of nist}
\end{center}
\end{table}

\subsection{Results}\label{results}

From Table~\ref{accuracy of copy} and \ref{bleu results of nist} that show Chinese-English experiment results , we can see that the baseline Transformer model can correctly translate 74\% of the sentences and 85\% of the NEs. 

The replacement method can increase the success rate at the sentence level to about 93\%, which is a notable improvement. 
However, it is when the tagging method is combined with other methods that the best performance is obtained. 
The best accuracy at the phrase level, of 99.15\%, is seen when combining tagging either with the replacement or the extra embeddings methods.
And the best accuracy at the sentence level, of 98.40\%, is obtained when the three proposed methods are combined together.

In terms of BLEU, with a score of 47.06, that is 1.4 points better than the base Transformer model, the combination of the three proposed methods is also the top scoring experiment under this metric.

\begin{table}
\begin{center}
\caption{Accuracy and BLEU scores of English-German translation experiments. Sentence and Phrase represent accuracy of translating pre-specfied translations at the sentence and phrase level. $BLEU_{all}$ and $BLEU_{*}$ denote BLEU scores on the whole test set and on sentences bear at least one replacement of an NE.}
\begin{tabular}{@{}l|llll@{}}
\toprule
        & Sentence & Phrase & $BLEU_{all}$ & $BLEU_*$ \\ \midrule
BASE    & 91.18\%    & 93.69\%      & 33.17      & 34.35      \\ \midrule
+T\&M\&E & \textbf{99.34\%}   & \textbf{99.54\%}      & \textbf{33.23}      & \textbf{34.71}      \\ \bottomrule
\end{tabular}
\label{table-ende}
\end{center}
\end{table}

From Table \ref{table-ende}, we also observe a significant improvement of accuracy on English-German translation. The phrase-level accuracy is increased by almost 6 points. If we compute BLEU scores on sentences containing NEs, our model is 0.36 BLEU points better than the baseline. In comparison to Chinese-English, the improvement of the BLEU score on English-German is not that big. The reasons are two-fold. First, the phrase-level accuracy of the baseline is already as high as 93.69\% since English and German share a large amount of NEs in the dataset.  Second, sentences containing NEs are less than 1/3 in the English-German test set.

\section{Analysis}\label{4}

In a more detailed analysis, we offer the following observations.
The tagging method has a weak impact when used alone, but there is a good improvement when it is combined with the replacement method. The reason for this seems to be that the source-side and the target-side tags in the tagging method help to establish a stronger connection when the source and target embeddings are shared, which is made possible with the replacement method.
  
The extra embeddings method and the tagging method seem both to enhance the copy signal, but the latter has a better impact than the former. The advantage of the extra embeddings method, nevertheless, is that it offers a softer approach, without requiring the modification of the data: this modification increases the length of the sentence and the risk of degrading the NMT performance.
  
At the phrase level, the accuracy of the mixed method is almost identical to the simpler replacement method, where a source phrase is directly replaced with its translation. But the mixed method leads to a higher BLEU score, thus retaining more  information relevant for translation.

We further conducted an in-depth analysis to study how and why the proposed methods work.


\subsection{Analysis of Embedding and Attention}\label{4.1}


\textbf{Embedding}
When using the R or the M method, the shared word embeddings gradually become cross-lingual word embeddings. This can be observed by extracting the embeddings of the T\&M\&E model and by calculating the nearest neighbours among words with them, using cosine after normalization. 

Table~\ref{table of nearest neighbours} displays two examples, with the words ``india''/``印度'' and ``beijing''/``北京''. For instance, the nearest neighbour of ``india'' is ``印度'', its equivalent in Chinese, while the word ``india'' is also the fourth nearest neighbour of ``印度''.

The replacement phrases seem thus to work as anchor points by means of which the model can gradually learn cross-lingual information. From previous work \cite{lample2018phrase,artetxe2018iclr}, cross-lingual embeddings are known to be sometimes beneficial to translation quality.

\begin{figure}[t]
\centering
\small
\includegraphics[scale=0.65]{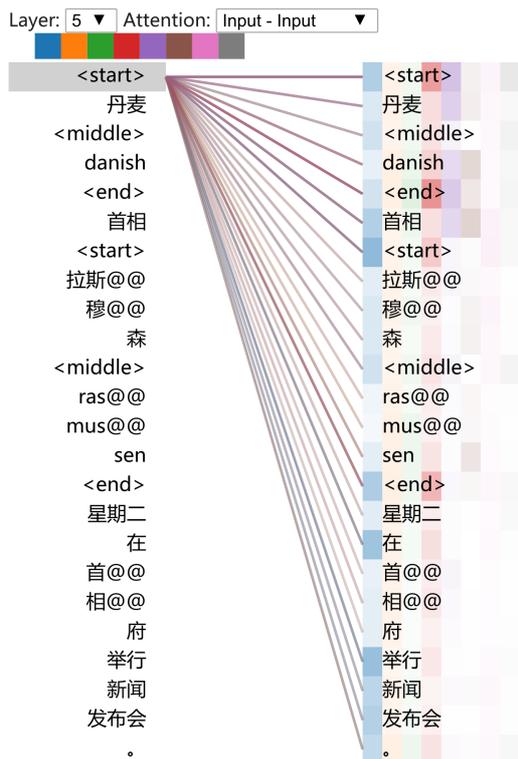}
\caption{An example of tagging pattern of T\&M\&E's model in the encoder self-attention in layer 5. The eight different colors represent the eight different heads. Here $<$start$>$ attends to other tags in the first and fourth attention heads.}
\label{self-attention figure}
\end{figure}
\noindent\textbf{Attention}
Transformer uses multi-heads to attend to information from different representation sub-spaces at different positions.


\begin{table}
\begin{center}
\caption{The 7 nearest neighbours of words ``india''/``印度'' and ``beijing''/``北京''.}
\begin{tabular}{@{}llll@{}}
\toprule
india   & 印度     & beijing    & 北京       \\ \midrule
印度      & 印      & 北京         & beijing  \\
印       & 印@@    & 京          & 京        \\
japan   & 印度@@   & 北京@@       & 北京@@     \\
russia  & india  & china      & 来京       \\
namibia & 印中     & washington & 台北       \\
印度@@    & 印方     & shanghai   & 京@@      \\
印中      & indian & 京@@        & johann@@ \\ \bottomrule
\end{tabular}
\label{table of nearest neighbours}
\end{center}
\end{table}

\begin{figure}[tt]
\centering
\includegraphics[scale=0.55]{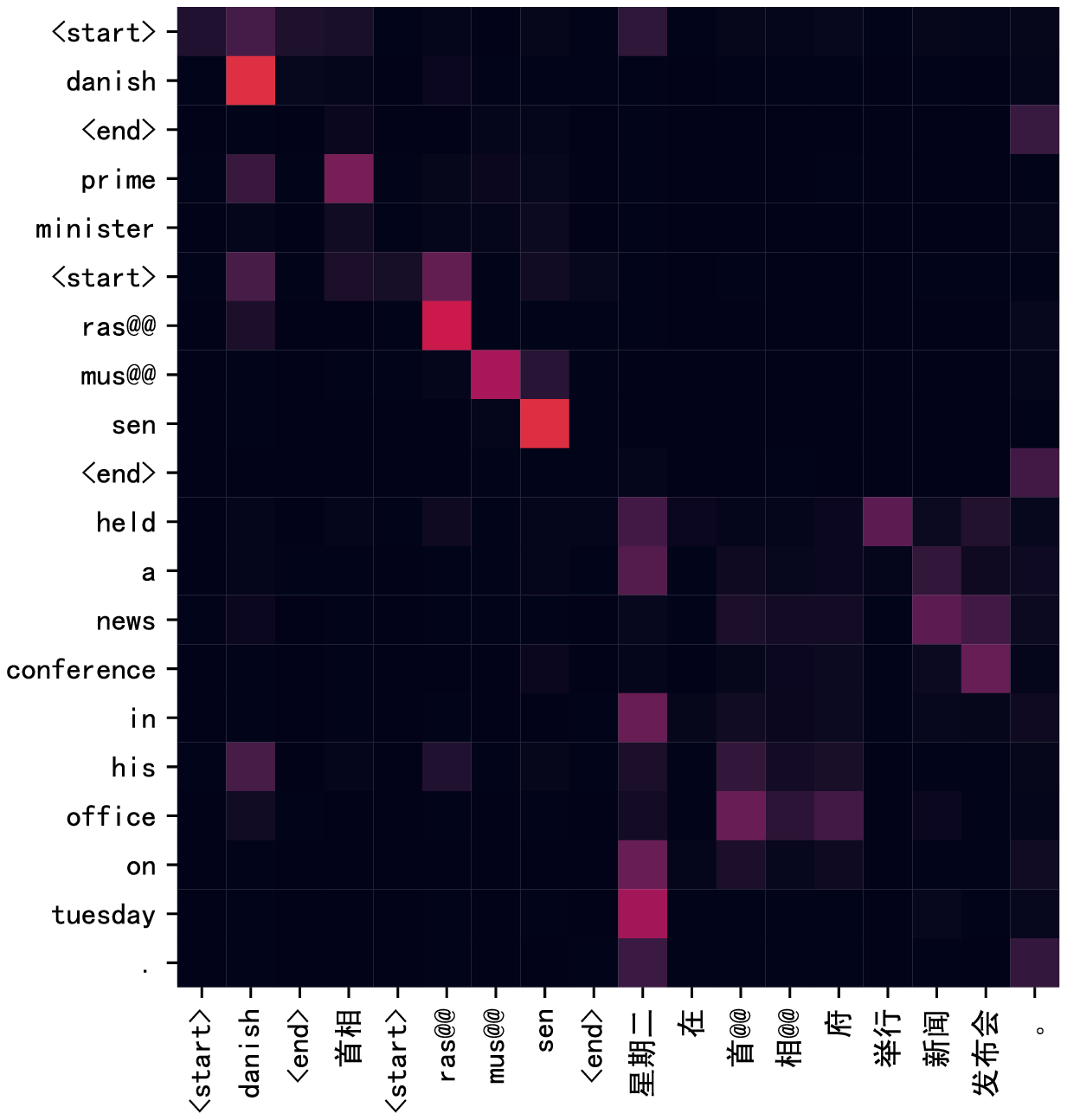}
\includegraphics[scale=0.5]{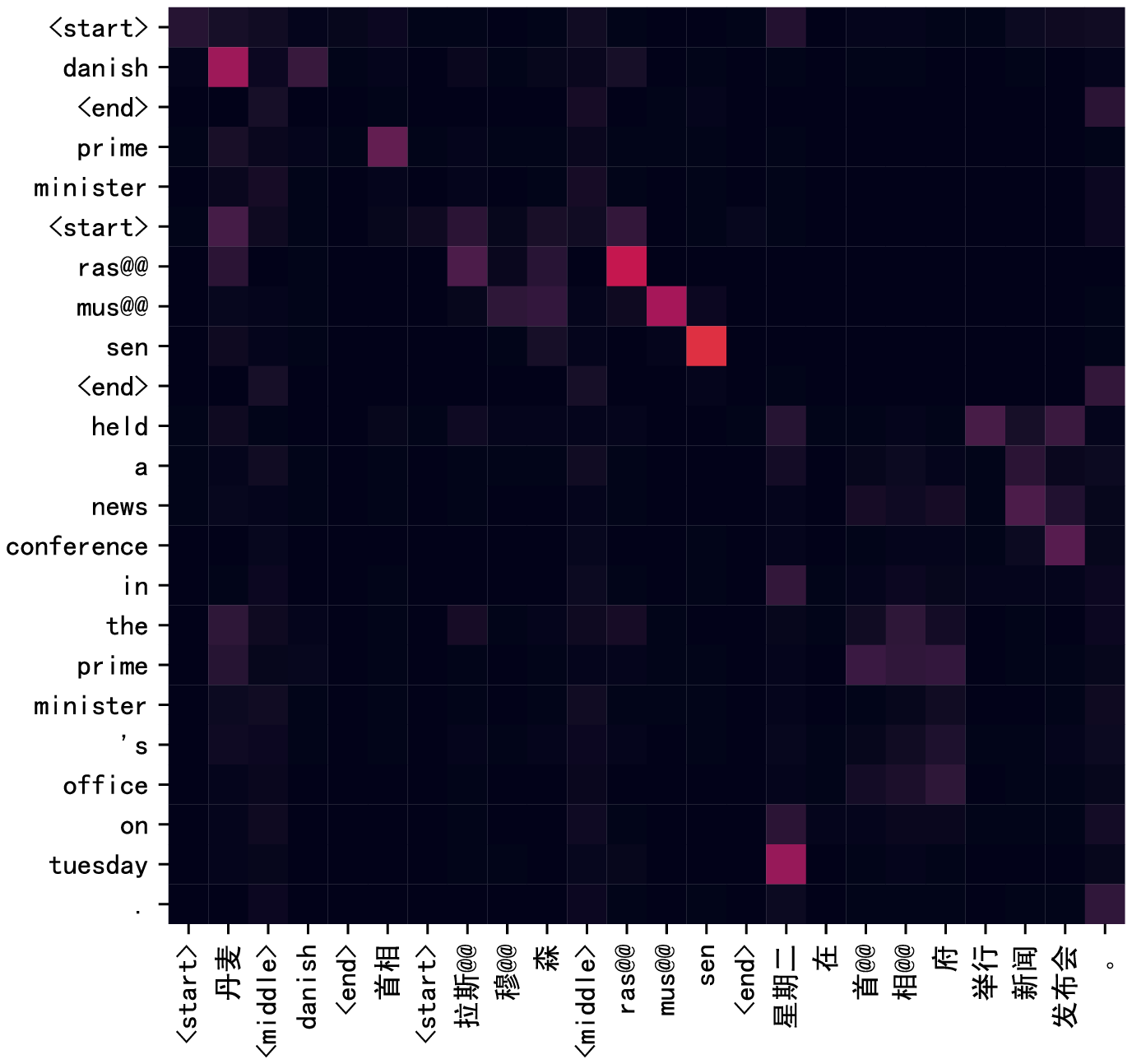}
\caption{Two examples of averaged cross attention in decoder last layer in T\&R (top) and T\&M\&E (bottom) models. The lighter the color, the higher the attention.}
\label{cross attention figure}
\end{figure}

An example of self-attention for the tag $<$\texttt{start}$>$ in T\&M\&E is displayed in Fig.~\ref{self-attention figure}. 
We can observe that it has strong connections with other tags in the first and fourth head.\footnote{
We don't use averaged self-attention to show how tagging works because after averaging, attention in a single head is hard to recognize. This is why we cannot see obvious connection of tags in cross-attention in Fig.~\ref{cross attention figure}.}
This is the type of attention pattern that is expected to help with the translation of the NEs enclosed in the tags through the tagging methods.

Examples of the attention matrix are, in turn, shown in Fig.~\ref{cross attention figure}. In the top matrix, for a T\&R model, one can observe that during decoding, the expressions ``danish'' and ``ras@@ mus@@ sen'' in the target side have high attention to the respective source-side parts in the tagged phrases. 

In the bottom matrix, now for a T\&M\&E model, the target-side phrase has an attention to the respective source phrase part and to the to-copy part at the same time. This vividly illustrates the role of the mixed phrase method. 
Moreover, it is worth noting that ``danish'' is directly translated from ``丹麦'' (danish), as the model is very certain of the answer. When translating ``ras@@ mus@@ sen'', because the number of occurrences of such expressions is comparatively small, the certainty of the translation is not high enough and the to-copy part in the source-side gets more attention.

\subsection{Error Analysis}\label{4.2}

The application of the methods proposed to handle NEs, in particular when the tagging method is combined with the replacement or with the mixed methods (T\&R or T\&M), increases the success rate of translation at the phrase level to more than 99\%.
As indicated in Table~\ref{accuracy of copy}, there are still about 50 NEs that nevertheless were not successfully translated. We report now on the manual verification of these errors and the lessons learned.

\begin{table*}
\begin{center}
\caption{Two examples of errors involving the T\&R method. We don’t show the entire sentences with all words due to the space limit in the second example.}
\begin{tabular}{cl}
\toprule
T\&R SRC & \begin{tabular}[c]{@{}l@{}}这 起 诉讼 由 六 名 中国人 提出 , 他们 在 二 次 大战 期间 被迫 在 \textless{}start\textgreater\ japanese \textless{}end\textgreater \\  一 座 镍 矿山 作 奴@@ 工 。\end{tabular}                                                                                                                                                       \\ \midrule
T\&R OUT & \begin{tabular}[c]{@{}l@{}}the lawsuit was proposed by six chinese who were forced to work in a \textless{}\ start\textgreater\ japanese \textless{}end\ \textgreater\ \\ nickel mine during world war ii .\end{tabular}                                                                                                             \\ \midrule
REF      &  \begin{tabular}[c]{@{}l@{}}lawsuit filed by six chinese who were forced to work as slave laborers in a nickel mine in japan \\ during world war ii .\end{tabular}                                                                                                                                      \\ \midrule
T\&R SRC & \begin{tabular}[c]{@{}l@{}}这个 总部 设 在 \textless{}start\textgreater\ in \textless{}end\textgreater\ 的 环境 研究 组织 也 称@@ 许 地方性 的 环保 成就 , 例\\如 \textless{}start\textgreater\ the netherlands \textless{}end\textgreater\ 创下 百分之八@@ 十六 的 汽车 回收率;\end{tabular}                                                                             \\ \midrule
T\&R OUT & \begin{tabular}[c]{@{}l@{}}\textless{}start\textgreater\ the netherlands \textless{}end\textgreater\ - based environmental research institute also commended denmark\\ for its achievements \textless{}start\textgreater\ in \textless{}end\textgreater ...\end{tabular} \\ \midrule
REF      & \begin{tabular}[c]{@{}l@{}}this washington-based environmental research organization also praised local successes in\\ environmental protection. for example , the netherlands has achieved an 86 percent recycling rate\\ for automobiles ;\end{tabular} \\ \bottomrule            
\end{tabular}
\label{table of error samples}
\end{center}
\end{table*}

A first aspect to note is that among the 50 errors, about 15 occur with other errors in the same sentences, which indicate some commonalities among them. 
A large part of the errors has its root in the construction of the bilingual database and in the pre-processing of the data with that database. It is inevitable that some noise is introduced especially when for a given source phrase there may be multiple target phrases, and only one of them was stored in the database to be retrieved during pre-processing. In some sentences, it may be the target options that were left out of the bilingual database that would be the correct translation.

This is illustrated in the first example of Table~\ref{table of error samples}. ``日本'' is paired with ``japanese'' in the external database and replaced by it during data pre-processing, but it is ``japan'' that would have been the reference replacement. Observing the corpus in more detail, one notes that in the sentences before this example, ``日本'' appears many times, and in most of them its translation is ``japanese'', which was eventually stored in the database, and not ``japan'', due to its higher frequency.

Interestingly, the translation of ``日本'' by ``japanese'' eventually contributed to a target sentence that is also a correct translation although not consistent with the reference translation. This kind of mismatch is common when translating from Chinese. For example, ``克罗地亚总理'' and ``克罗地亚'' have the same sub-phrase ``克罗地亚'', but in English they are translated into ``croatian prime minister'', and into ``croatia''.

The T\&R method is specially vulnerable to this kind of replacement errors during pre-processing. An error of this type may result in a subsequent error in the translation of the phrase affected by it (as in the example above) or even in the entire sentence being ignored, as illustrated in the second example in Table~\ref{table of error samples}. Due to a replacement error --- ``华盛顿'' being replaced by ``in'' ---, the model copies ``in'' into the target sentence, skips the first source sentence and translates its left content. 

Another type of error occurs when the signal was not strong enough for the T\&M model to apply the replacement. One example of this type of errors can be found with the translation of ``罗斯@@ 尤可夫''. 
It should have been translated as ``los@@ yukov'', but the translation model also considered what it had learned about the sub-phrase ``罗斯'' outside this specific context, and eventually chose it, delivering the translation ``ross@@ yukov''. This situation occurs many times, with the system choosing to ``translate'' according to general information learned rather than to ``copy'' according to specific information due to tagging methods.


\section{Related Work}


Methods aiming at incorporating external translations into NMT can be roughly grouped into two clusters: 1) translating out-of-vocabulary (OOVs) or low-frequency words with external dictionaries; and 2) incorporating phrase translations from a Statistical Machine Translation (SMT) model or from a bilingual dictionary into NMT by changing either the NMT model or the decoding algorithm. 

In the first type of methods, Luong et al.~\cite{luong2015addressing} introduce an alignment-based technique that trains NMT systems on data that are augmented by the output of a word alignment algorithm and then the OOVs are translated by using a dictionary in a post-processing step. Arthur et al.~\cite{arthur2016incorporating} incorporate discrete translation lexicons into NMT to translate low-frequency words. Zhang et al.~\cite{Zhang2016BridgingNM} propose two methods to bridge NMT and bilingual dictionaries. Most of the rare words in the test sentences can obtain correct translations if they are covered by the bilingual dictionary in the training process. The proposals above focus mainly on word-level OOV translation, with a limited exploration of multi-word phrases for NMT.

In the second type of approaches, various methods have  been proposed to equip NMT with the ability of translating multi-word phrases that are either from an SMT model or from a bilingual dictionary. Stahlberg et al.~\cite{stahlberg2016syntactically} use the phrase translations produced by a hierarchical phrase-based SMT system as hard decoding constraints for the NMT decoder in order to enable NMT to generate more syntactic phrases. 
Tang et al.~\cite{tang2016neural} propose phraseNet to make NMT decoder generate phrase translations according to an external phrase memory. Wang et al.~\cite{wang2017translating} introduce a method to translate phrases in NMT by integrating a phrase memory that stores target phrases from a phrase-based SMT (PBSMT) model into the encoder-decoder framework of NMT. Zhang et al.~\cite{zhang2017prior} represent prior knowledge sources as features in a loglinear model and propose to use posterior regularization to provide a general framework for integrating prior knowledge into NMT. Dahlmann et al.~\cite{dahlmann2017neural} use a log-linear model to combine the strengths of NMT and PBSMT models and propose a novel hybrid search algorithm that is extended with phrase translations from SMT. These efforts focus on allowing NMT to translate phrases by changing the NMT model or applying loglinear models.

Song et al.~\cite{song2019code}, in turn, produce code-switched training data by replacing source phrases with their target translations and use a pointer network to enhance their copy into the target side. This method is similar to our mixed phrase method.
Unfortunately, they cannot guarantee that external phrase translations are present in the final translations generated by the augmented NMT model since this is not their goal.

Yet another approach to incorporate external phrase translations into NMT is to change the beam search algorithm in the decoder. Hokamp et al.~\cite{hokamp2017lexically} propose a Grid Beam Search algorithm that allows specific sub-sequences to be present in the output of the model. The sub-sequence can be either single- or multi-word expressions. Chatterjee et al.~\cite{chatterjee2017guiding} further propose a “guiding” mechanism that enhances an existing NMT decoder with the ability to handle translation recommendations in the form of XML annotations of source words (e.g. terminology lists). 

The two approaches just mentioned above incorporate external knowledge into NMT with enhanced beam search algorithms. Although they do not change NMT architectures, they have to decide at each time step of the decoding process whether to find a translation option from translation recommendations in the form of a bilingual lexicon or phrase table, or to generate a target word according to the NMT models. This degrades the decoding speed of NMT.

When compared to previous work, our approach is characterized by its combined strengths of (i) its ability to handle both word- and phrase-level pre-defined translations; (ii) endowing NMT with a very high success rate, of over 99\%, as presented in Section \ref{results}; (iii) and no degradation in terms of decoding speed. Additionally, since we only add extra embeddings to the network architecture, our approach is easily reproduced.

\section{Conclusion and Future Work}

In this paper, we propose the TME approach to merge bilingual pairs into NMT. This is a combination of three methods to endow NMT with pre-specified translations. In a pre-processing stage, the beginning and the end of textual occurrences of expressions from an external database are marked with special purpose tags (T: tagging method). Additionally, the source expressions are added with their translational equivalent target expressions (M: mixed phrases method). And an extra embedding is also applied to distinguish pre-specified phrases from other words (E: extra embeddings method).

Results from experimentation and further analysis indicate that the tagging method helps that a type of pattern is learned through the attention mechanism and that this is enhancing the copy signal; that the shared embeddings gradually learn cross-lingual information with the help of the additional target phrases as anchor points, and that this is helping the translation of pre-specified phrases; and that using extra embeddings is a softer approach when compared with tagging, though its impact is not as good, but yet that it is able to improve the quality of the final translation.

Experiments in Chinese-English and English-German translation in the news domain confirm the effectiveness of the approach proposed here, with a substantial increase in translation quality and accuracy, with more than 99\% of the pre-specified phrases being correctly translated.

In the future, we would like to extend this approach from the translation of named entities to other areas, including other types of fixed phrases. Incorporating entities with multiple candidate translations could be one of interesting future research directions.


\bibliography{ecai}

\begin{thebibliography}{10}

\bibitem{artetxe2018iclr}
Mikel Artetxe, Gorka Labaka, Eneko Agirre, and Kyunghyun Cho, `Unsupervised
  neural machine translation', in {\em Proceedings of the Sixth International
  Conference on Learning Representations}, (April 2018).

\bibitem{arthur2016incorporating}
Philip Arthur, Graham Neubig, and Satoshi Nakamura, `Incorporating discrete
  translation lexicons into neural machine translation', in {\em Proceedings of
  the 2016 Conference on Empirical Methods in Natural Language Processing}, pp.
  1557--1567, (2016).

\bibitem{chatterjee2017guiding}
Rajen Chatterjee, Matteo Negri, Marco Turchi, Marcello Federico, Lucia Specia,
  and Fr{\'e}d{\'e}ric Blain, `Guiding neural machine translation decoding with
  external knowledge', in {\em Proceedings of the Second Conference on Machine
  Translation}, pp. 157--168, (2017).

\bibitem{che2010ltp}
Wanxiang Che, Zhenghua Li, and Ting Liu, `Ltp: A chinese language technology
  platform', in {\em Proceedings of the 23rd International Conference on
  Computational Linguistics: Demonstrations}, pp. 13--16. Association for
  Computational Linguistics, (2010).

\bibitem{dahlmann2017neural}
Leonard Dahlmann, Evgeny Matusov, Pavel Petrushkov, and Shahram Khadivi,
  `Neural machine translation leveraging phrase-based models in a hybrid
  search', in {\em Proceedings of the 2017 Conference on Empirical Methods in
  Natural Language Processing}, pp. 1411--1420, (2017).

\bibitem{devlin2018bert}
Jacob Devlin, Ming-Wei Chang, Kenton Lee, and Kristina Toutanova, `Bert:
  Pre-training of deep bidirectional transformers for language understanding',
  {\em arXiv preprint arXiv:1810.04805}, (2018).

\bibitem{Hinton2012ImprovingNN}
Geoffrey~E. Hinton, Nitish Srivastava, Alex Krizhevsky, Ilya Sutskever, and
  Ruslan~R. Salakhutdinov, `Improving neural networks by preventing
  co-adaptation of feature detectors', {\em CoRR}, {\bf abs/1207.0580}, (2012).

\bibitem{hokamp2017lexically}
Chris Hokamp and Qun Liu, `Lexically constrained decoding for sequence
  generation using grid beam search', in {\em Proceedings of the 55th Annual
  Meeting of the Association for Computational Linguistics (Volume 1: Long
  Papers)}, pp. 1535--1546, (2017).

\bibitem{adam}
Diederik~P. Kingma and Jimmy Ba, `Adam: A method for stochastic optimization',
  in {\em Proceedings of ICLR}, (2015).

\bibitem{koehn2007moses}
Philipp Koehn, Hieu Hoang, Alexandra Birch, Chris Callison-Burch, Marcello
  Federico, Nicola Bertoldi, Brooke Cowan, Wade Shen, Christine Moran, Richard
  Zens, et~al., `Moses: Open source toolkit for statistical machine
  translation', in {\em Proceedings of the 45th annual meeting of the
  association for computational linguistics companion volume proceedings of the
  demo and poster sessions}, pp. 177--180, (2007).

\bibitem{lample2018phrase}
Guillaume Lample, Myle Ott, Alexis Conneau, Ludovic Denoyer, et~al.,
  `Phrase-based \& neural unsupervised machine translation', in {\em
  Proceedings of the 2018 Conference on Empirical Methods in Natural Language
  Processing}, pp. 5039--5049, (2018).

\bibitem{luong2015addressing}
Thang Luong, Ilya Sutskever, Quoc Le, Oriol Vinyals, and Wojciech Zaremba,
  `Addressing the rare word problem in neural machine translation', in {\em
  Proceedings of the 53rd Annual Meeting of the Association for Computational
  Linguistics and the 7th International Joint Conference on Natural Language
  Processing (Volume 1: Long Papers)}, volume~1, pp. 11--19, (2015).

\bibitem{nadeau2007survey}
David Nadeau and Satoshi Sekine, `A survey of named entity recognition and
  classification', {\em Lingvisticae Investigationes}, {\bf 30}(1),  3--26,
  (2007).

\bibitem{ott2018scaling}
Myle Ott, Sergey Edunov, David Grangier, and Michael Auli, `Scaling neural
  machine translation', in {\em Proceedings of the Third Conference on Machine
  Translation (WMT)}, (2018).

\bibitem{papineni2002bleu}
Kishore Papineni, Salim Roukos, Todd Ward, and Wei-Jing Zhu, `Bleu: a method
  for automatic evaluation of machine translation', in {\em Proceedings of the
  40th annual meeting on association for computational linguistics}, pp.
  311--318. Association for Computational Linguistics, (2002).

\bibitem{sennrich2016neural}
Rico Sennrich, Barry Haddow, and Alexandra Birch, `Neural machine translation
  of rare words with subword units', in {\em Proceedings of the 54th Annual
  Meeting of the Association for Computational Linguistics (Volume 1: Long
  Papers)}, volume~1, pp. 1715--1725, (2016).

\bibitem{song2019code}
Kai Song, Yue Zhang, Heng Yu, Weihua Luo, Kun Wang, and Min Zhang,
  `Code-switching for enhancing nmt with pre-specified translation', {\em arXiv
  preprint arXiv:1904.09107}, (2019).

\bibitem{stahlberg2016syntactically}
Felix Stahlberg, Eva Hasler, Aurelien Waite, and Bill Byrne, `Syntactically
  guided neural machine translation', in {\em The 54th Annual Meeting of the
  Association for Computational Linguistics}, p. 299, (2016).

\bibitem{szegedy2016rethinking}
Christian Szegedy, Vincent Vanhoucke, Sergey Ioffe, Jon Shlens, and Zbigniew
  Wojna, `Rethinking the inception architecture for computer vision', in {\em
  Proceedings of the IEEE conference on computer vision and pattern
  recognition}, pp. 2818--2826, (2016).

\bibitem{tang2016neural}
Yaohua Tang, Fandong Meng, Zhengdong Lu, Hang Li, and Philip~LH Yu, `Neural
  machine translation with external phrase memory', {\em arXiv preprint
  arXiv:1606.01792}, (2016).

\bibitem{vaswani2017attention}
Ashish Vaswani, Noam Shazeer, Niki Parmar, Jakob Uszkoreit, Llion Jones,
  Aidan~N Gomez, {\L}ukasz Kaiser, and Illia Polosukhin, `Attention is all you
  need', in {\em Advances in neural information processing systems}, pp.
  5998--6008, (2017).

\bibitem{wang2017translating}
Xing Wang, Zhaopeng Tu, Deyi Xiong, and Min Zhang, `Translating phrases in
  neural machine translation', in {\em Proceedings of the 2017 Conference on
  Empirical Methods in Natural Language Processing}, pp. 1421--1431, (2017).

\bibitem{zhang2017prior}
Jiacheng Zhang, Yang Liu, Huanbo Luan, Jingfang Xu, and Maosong Sun, `Prior
  knowledge integration for neural machine translation using posterior
  regularization', in {\em Proceedings of the 55th Annual Meeting of the
  Association for Computational Linguistics (Volume 1: Long Papers)}, pp.
  1514--1523, (2017).

\bibitem{Zhang2016BridgingNM}
Jiajun Zhang and Chengqing Zong, `Bridging neural machine translation and
  bilingual dictionaries', {\em CoRR}, {\bf abs/1610.07272}, (2016).

\end{thebibliography}
\end{CJK}
\end{document}